\documentclass[twoside]{article}

\usepackage[accepted]{aistats2023}
\usepackage{graphicx}
\usepackage{booktabs}
\usepackage{tabularx}
\usepackage{wrapfig}
\usepackage[dvipsnames]{xcolor}
\usepackage{tikz}
\usepackage{comment}
\usepackage{amsmath,amssymb}
\usepackage{color}
\usepackage{subcaption}

\DeclareMathOperator*{\argmax}{arg\,max}

\newcommand{\NNnote}[1]{\textcolor{red}{{{Nikhil:} #1}}}

\newcommand{\Rise}[1]{\textcolor{Green}{\textsubscript{\bf $+$#1}}}

\makeatletter
\newcommand\notsosmall{\@setfontsize\notsosmall\@viiipt{10.8}}
\makeatother
\usepackage[pagebackref,breaklinks,colorlinks]{hyperref}
\usepackage[capitalize]{cleveref}
\crefname{section}{Sec.}{Secs.}
\Crefname{section}{Section}{Sections}
\Crefname{table}{Table}{Tables}
\crefname{table}{Tab.}{Tabs.}

\usepackage[accsupp]{axessibility}  % Improves PDF readability for those with disabilities.
\usepackage[capitalize]{cleveref}

\usepackage[round]{natbib}

\newcommand{\EE}{\mathbb{E}}

\begin{document}
\twocolumn[

\aistatstitle{CLIP-Lite: Information Efficient Visual Representation Learning with Language Supervision}

\aistatsauthor{ Aman Shrivastava \And Ramprasaath R. Selvaraju \And  Nikhil Naik \And Vicente Ordonez }

\aistatsaddress{ University of Virginia \And  Salesforce Research \And Salesforce Research \And Rice University } ]

%%%%%%%%% BODY TEXT
\begin{abstract}
We propose CLIP-Lite, an information efficient method for visual representation learning by feature alignment with textual annotations. Compared to the previously proposed CLIP model, CLIP-Lite requires only one negative image-text sample pair for every positive image-text sample during the optimization of its contrastive learning objective. We accomplish this by taking advantage of an information efficient lower-bound to maximize the mutual information between the two input modalities. This allows CLIP-Lite to be trained with significantly reduced amounts of data and batch sizes while obtaining better performance than CLIP at the same scale. We evaluate CLIP-Lite by pretraining on the COCO-Captions dataset and testing transfer learning to other datasets. CLIP-Lite obtains a +14.0\% mAP absolute gain in performance on Pascal VOC classification, and a +22.1\% top-1 accuracy gain on ImageNet, while being comparable or superior to other, more complex, text-supervised models. CLIP-Lite is also superior to CLIP on image and text retrieval, zero-shot classification, and visual grounding. Finally, we show that CLIP-Lite can leverage language semantics to encourage bias-free visual representations that can be used in downstream tasks. Implementation: \url{https://github.com/4m4n5/CLIP-Lite}
\end{abstract}

\section{Introduction}
\label{sec:introduction}
\begin{figure}[t]
    \centering
    \includegraphics[width=1.0\linewidth]{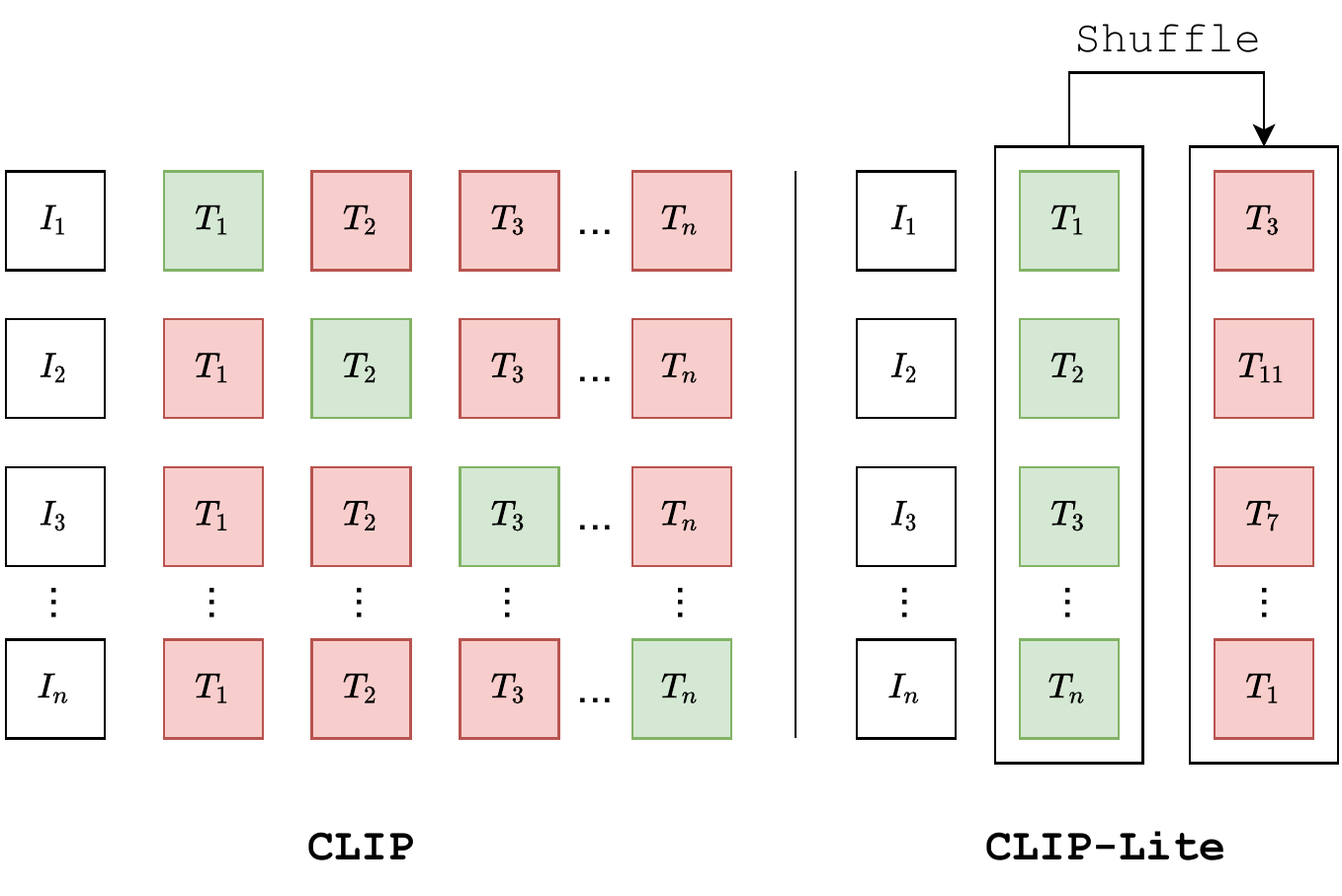}
    % \vspace{0.0in}
    \caption{Given a batch of $n$ image-caption pairs $\{(I_i,T_i)\}$, CLIP requires a large number of negative pairs $\{(I_i,T_j) \mid i \neq j\}$ due to the need to pair every image in the batch with captions from other images. Whereas, CLIP-Lite can learn representations using a single negative pair (in red) for every positive pair (in green).
    }%, reducing the batch size to $O(n)$.}
    \label{fig:teaser}
\end{figure}

Pretraining image classification networks on the Imagenet dataset has led to visual representations that transfer to other tasks~\citep{girshick2014rich,long2015fully,vinyals2015show,antol2015vqa,zhu2016visual7w}. However, such classification based pretraining requires a large amount of human-annotated data which is hard to obtain at scale. In contrast, captioned image data is an information-dense source of supervision that is relatively cheap to collect and plentiful on the internet. Therefore, recent methods have used joint vision-language pretraining to learn representations from image-caption pairs~\citep{desai2021virtex,sariyildiz2020learning}. However, methods such as VirTex~\citep{desai2021virtex} which train on complex language modeling tasks such as masked language modeling, token classification, and captioning fail to align features in a common latent space.

Recently, CLIP~\citep{radford2021learning}, a vision-language pretraining model, was developed using contrastive learning between the two modalities on an Internet-sized dataset of 400 million image-caption pairs. Contrastive learning methods work by pulling closer the representations of independent views of the same datum \textit{i.e.}~a positive or matching image-caption pair and pushing apart the representations of independent views of different data \textit{i.e.}~negative or non-matching image-caption pairs. However, contrastive learning in vision-language pretraining still has some limitations as it seems to be most effective only with large scale data, and it requires a large number of negative image-caption pairs during training. Our work aims to address and explore these two limitations by proposing CLIP-Lite, an information efficient variation of CLIP that is useful even in smaller data regimes, does not rely in as many negative sample pairs during training, and provides comparable or superior performance on standard benchmarks against other methods trained at the same scale. Our work is motivated by the observation that multiple contrastive objectives maximize a lower-bound on the mutual information between two or more views of the same datum~\citep{wu2020mutual}. 

CLIP particularly maximizes the mutual information between the image and its caption by using a mutual information lower bound based on InfoNCE~\citep{oord2018representation}. The InfoNCE bound has seen wide adoption due to its favorable properties such as stability and low variance. However, the the bound is theoretically loose in cases when the true mutual information is larger than $\log K$ where $(K-1)$ is the number of negative samples used for training. The negative pairs can be randomly sampled but usually a large amount of negative pairs are required to have a good estimate of the mutual information between the two input streams, and hence the need for rather large batch sizes~\citep{bachman2019learning,chen2015microsoft} or memory-banks~\citep{chen2020improved,tian2019contrastive,he2020momentum}. 

We instead adopt a lower bound based on Jenssen Shannon Divergence to maximize the mutual information~\citep{hjelm2018learning,nowozin2016f}, thus requiring no more than one negative example pair for each positive example pair. This reduces the number of negative examples in a training batch to $O(n)$, where $n$ is the batch size. In contrast, CLIP uses $O(n^2)$ negative example pairs per batch. Figure~\ref{fig:teaser} illustrates this difference.  We implement this strategy and demonstrate thoroughly the efficacy of CLIP-Lite through experiments on several tasks and datasets at various scales. Our method demonstrates impressive data efficiency and is able to outperform CLIP trained on the entire COCO-Captions dataset while only training on $20\%$ of the same dataset. We also demonstrate that CLIP-Lite can be used as a good source of pretrained features by showing good generalization on Pascal VOC and Imagenet classification. We also show that the visual feature backbone of CLIP-Lite can be finetuned in the iNaturalist dataset to match top performances on this benchmark with caption supervision pretraining. Furthermore, we show that CLIP-Lite leads to good visual features for image retrieval compared to regular CLIP trained on COCO Captions. We also demonstrate that CLIP-Lite enables the removal of concepts from visual representations which we show can be applied in bias mitigation. Our work extends and complements the work using contrastive learning, especially addressing the computational requirements of the original CLIP model in terms of memory overhead through minimizing the number of negative sample image-text pairs required during training and shows its effectiveness in smaller data regimes including for zero-shot learning on CIFAR-10, image-text retrieval and unsupervised object localization.

\section{Related Work}
\label{sec:background}
Our work is related to several strands of research on visual pretraining without full-supervision.

\noindent\textbf{Vision-Language Pretraining:} Research on learning visual representations by using textual labels or annotations has a long history. In \citep{quattoni2007learning}, the authors learn data-efficient image representations using manifold learning in the weight space of classifiers trained to predict tokens in image captions. Following this work, \citep{joulin2016learning} used convolutional neural networks to predict words in image captions to learn image representations. 

This approach was later extended in~\citep{lei2015predicting} where the model learns to predict phrase n-grams, which demonstrated impressive zero-shot performance on downstream classification tasks. 
Recently, VirTex~\citep{desai2021virtex} used proxy language modeling tasks, such as image-captioning to train a visual encoder and a transformer based language decoder which generates captions. ICMLM~\citep{sariyildiz2020learning} demonstrated a similar masked language modeling approach but relied on pretrained textual encoders for generating textual features. In ~\citep{stroud2020learning}, video representations are learned using paired textual metadata, however the method does not extend to visual pretraining for images.  In general, these methods distill the rich semantic information from a caption into the visual representation by learning to predict each token in the caption given the corresponding image. More recent work, such as CLIP~\citep{radford2021learning}, has shown that a simpler contrastive objective for aligning image and caption pairs is also able to learn a powerful visual representation. Our work extends CLIP using a more information-efficient approach. 

\noindent\textbf{Contrastive Representation Learning and Mutual Information Estimation:} As demonstrated in~\citep{wu2020mutual}, we observe that contrastive frameworks learn by maximizing the mutual information (MI) between different views of a given data point. For images, this is achieved by maximizing the MI between different augmentations of the data as in SimCLR~\citep{chen2020simple,bachman2019learning}. While for sequential data such as conversational text, consecutive utterances can be considered as different views~\citep{stratos2018mutual}. Similarly, several other contrastive frameworks have been proposed that learn representations in domains such as images~\citep{grill2020bootstrap,caron2020unsupervised}, text~\citep{mikolov2013efficient,stratos2018mutual}, graphs~\citep{velivckovic2018deep}, and videos~\citep{jabri2020space}. 
The value of mutual information is extremely challenging to estimate, especially for the high-dimensional continuous representations used in deep learning. To this end, various tractable lower-bounds on mutual information are used for optimization. Recently, MINE~\citep{belghazi2018mutual} proposed a general-purpose parameterized neural estimator of mutual information. It uses a Donsker-Varadhan~\citep{donsker1983asymptotic} representation of KL-divergence as the lower-bound on mutual information. MINE~\citep{belghazi2018mutual} used a neural network critic to distinguish positive and negative pairs of samples. Another popular bound on mutual information that has seen wide adoption due to its low variance is the InfoNCE~\citep{oord2018representation} bound. In~\citep{hjelm2018learning}, the infoNCE bound on the mutual information is used for unsupervised representation learning. While it is used by several other methods for self-supervised~\citep{chen2020simple} representation learning for images. The capacity of the bound is limited by the number of contrastive samples used~\citep{mcallester2020formal}. Additionally, InfoNCE can underestimate large amounts of true MI which is generally the case with high-dimensional representations of natural images. To this end, DeepInfoMax~\citep{hjelm2018learning} proposed using a lower-bound on mutual information that is based on the Jensen-Shannon Divergence (JSD) instead of the traditional KL-divergence (KLD). The authors show that the JSD based lower bound is stable, differentiable, and can be optimized with just one negative sample. Inspired by this, we extend the use of this bound for vision-language pretraining and demonstrate its effectiveness through extensive experimental evaluations.

\section{CLIP-Lite}
\label{sec:method}
\begin{figure}[t]
    \centering
    \includegraphics[width=1.0\linewidth]{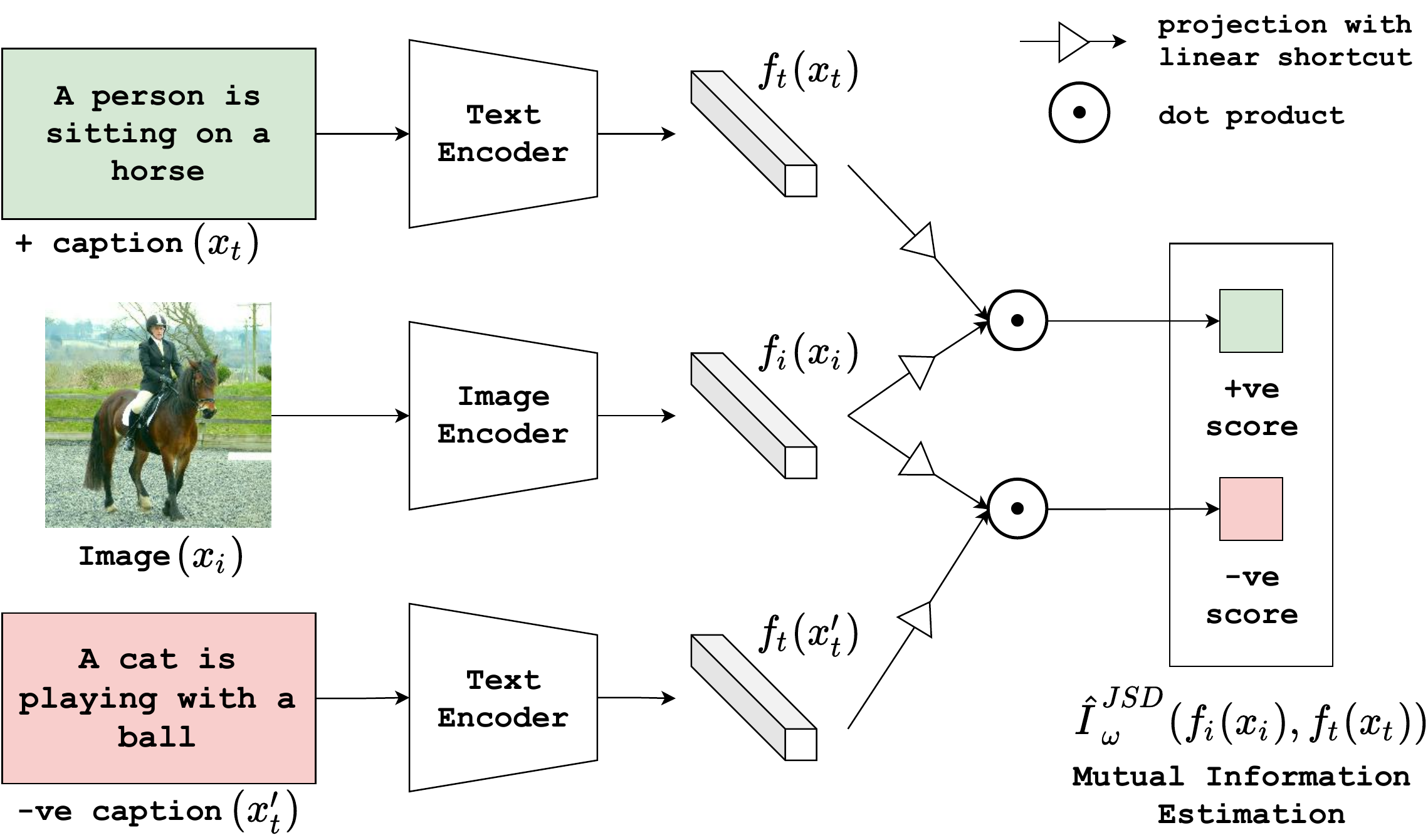}
    % \vspace{0.1in}
    \caption{\textbf{CLIP-Lite:} We extract representations for an image, its positive caption, and one negative caption. Image-caption pairs are then fed into the mutual information discriminator function which outputs a score for each pair. These scores are then used to estimate and maximize mutual information using Jensen-Shannon Divergence (JSD) to optimize the parameters of the encoders and the mutual information discriminator end-to-end. The projection and dot function represents the MI discriminator function $T_{\omega}$.
    }%, reducing the batch size to $O(n)$.}
    \label{fig:leadfigure}
    % \vspace{-0.3in}
\end{figure}

In this section, we describe our pretraining framework (Figure \ref{fig:leadfigure}) for visual representation learning. Given a dataset of image-caption pairs, the goal of our pretraining framework is to train an image encoder and a text encoder such that representations learned from the visual and the textual streams share maximum information (Figure \ref{fig:leadfigure} shows an overview). Consider an image encoder network, $f_i$ with parameters $\theta_i$ and a textual encoder, $f_t$ with parameters $\theta_t$. Let $(x_i, x_t)$ be a sampled image-caption pair from the dataset and $f_i(x_i)$ and $f_t(x_t)$ denote the representations extracted from the networks.  Based on the information bottleneck principle~\citep{tishby2015deep}, the maximum mutual information (MI) predictive coding framework~\citep{oord2018representation,hjelm2018learning,mcallester2020formal} aims to learn representations that maximize the MI between inputs and representations. In recent years, several methods~\citep{chen2020simple,he2020momentum,bachman2019learning} have used this principle to maximize MI between representations extracted from multiple views of a shared context. In the case of visual self-supervised learning, this is achieved by creating two independently-augmented copies of the same input and maximizing the MI between the respective features produced by an encoder. This framework can be extended further by considering an image $x_i$ and its caption $x_t$ as distinct views of the same input. This setup is motivated by the observation that image captions contain rich semantic information about images, for instance, presence of objects, location of objects, their relative spatial configurations, etc. Distilling this information into our visual representation is useful for robust representation learning~\citep{radford2021learning}. To this end, we formulate our objective as follows:
% \vspace{-0.1in}
\begin{equation}
    (\hat{\theta_i}, \hat{\theta_t}) = \argmax_{\theta_i, \theta_t} \; I(f_i(x_i), f_t(x_t)),
% \vspace{-0.13in}
\end{equation}

where $I(f_i(x_i), f_t(x_t)) \leq I(x_i; x_t);$ due to the data processing inequality between visual and textual streams.

% Subsection 
\subsection{Mutual Information Maximization} \label{sec:mim}
For given random variables $y$ and $z$, their mutual information is defined as a Kullback-Leibler (KL) divergence between their joint distribution $p(y,z)$ and the product of their marginal distributions, $p(y)p(z)$ as, 
\begin{equation} \label{eq:2}
    I(y; z) = D_{\text{KL}}(p(y, z) \, || \, p(y)p(z)).
\end{equation}
However, mutual information is notoriously hard to estimate for high-dimensional continuous variables, especially when the distributions $p(y,z)$, $p(x)$, or $p(z)$ are not explicitly known. As a result, recent approaches use various tractable lower bounds on the mutual information which are differentiable and hence can be maximized with gradient-descent based optimization. For contrastive learning, a commonly used bound is infoNCE~\citep{oord2018representation} based on Noise-Contrastive Estimation~\citep{gutmann2010noise}. This bound is relatively more stable and has been shown to work in a wide variety of tasks~\citep{chen2020simple,bachman2019learning,chen2020improved} including CLIP~\citep{radford2021learning} which, similar to our method, aims to learn visual representations from textual annotations. The infoNCE bound has seen wider adoption as it demonstrates lower variance compared to the Donsker-Varadhan bound~\citep{donsker1983asymptotic}. However, both of these bounds require a large number of negative samples and as a result, recent methods either train with extremely large batch-sizes~\citep{radford2021learning,chen2020simple}; or an additional memory-bank of negative samples~\citep{chen2020improved,tian2020makes}.

Unlike these works, we estimate mutual information using a Jensen-Shannon Divergence (JSD) bound, similar to formulations used for generative modeling~\citep{nowozin2016f}; and source separation~\citep{brakel2017learning}. This bound on mutual information is derived by replacing the KL-divergence in equation \ref{eq:2} with the Jensen-Shannon divergence (ref. appendix for further discussion). Interestingly, the lower bound derived as such is stable, differentiable, monotonically related to the mutual information $I(y; z)$, and most importantly, not dependent on the number of negative samples. Hence we have, $I(Y;Z) \geq \hat{I}_{\omega}^{JSD}(Y;Z)$ where,

\begin{equation} \label{eq:3}
\begin{split}
\hat{I}_{\omega}^{JSD}(Y;Z) := &\mathbb{E}_{P(Y, Z)}[-\mathrm{log}(1 + e^{-T_{\omega}})] \\ &- \mathbb{E}_{P(Y)P(Z)}[\mathrm{log}(1 + e^{T_{\omega}})],
\end{split}
\end{equation}

and $T_{\omega}: \mathcal{Y} \times \mathcal{Z} \to \mathbb{R}$ is a discriminator neural network with trainable parameters $\omega$ which are jointly optimized to distinguish between a paired-sample from a joint distribution (positive image-caption pair) and one pair from the product of marginals (negative image-caption pair). Therefore we are able to optimize our overall objective with just one negative sample as follows: 
\begin{equation}
    (\hat{\omega}, \hat{\theta_i}, \hat{\theta_t}) = \underset{\omega, \theta_i, \theta_t}{\mathrm{argmax}} \;\hat{I}^{JSD}_{\omega} (f_i(x_i), f_t(x_t)),
\end{equation}
where the visual encoder is a convolution neural network, and  features are extracted from the pre-classification layer of the network. The textual encoder is parameterized by a neural network that takes the caption as a string of textual-tokens and generates a one-dimensional representation.

\section{Experiments}
\label{sec:experiments}
In this section, we describe the experiments that demonstrate the value of using textual captions for learning visual representations using CLIP-Lite. In our experiments, the CLIP-Lite architecture consists of a ResNet-50 image encoder and the BERT-base textual encoder and is trained on the COCO Captions~\citep{chen2015microsoft} dataset. We evaluate the robustness of our visual encoder through the following downstream tasks which use the visual encoder (1) as a frozen feature extractor, or (2) as source of weight initialization for finetuning (ref. appendix). In addition, we also demonstrate the data efficiency of our method by evaluating performance on fractional datasets.

\begin{figure}[t]
    \centering
    \includegraphics[width=0.99\linewidth]{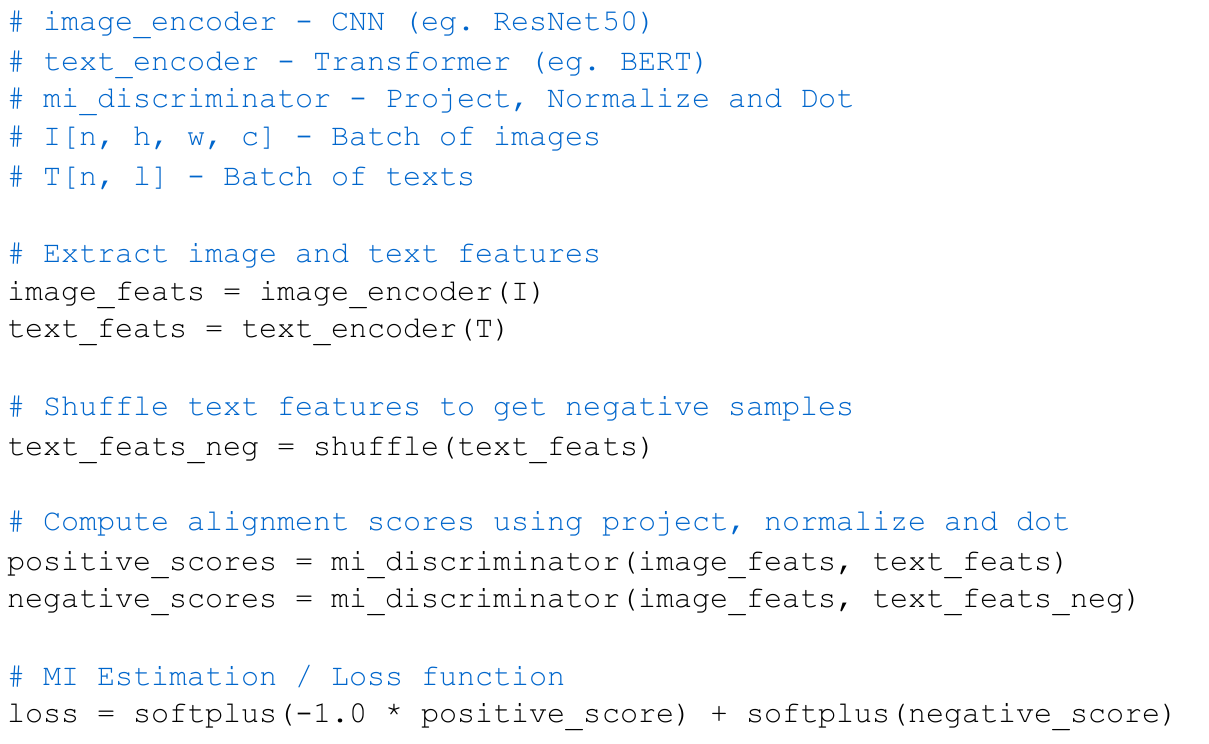}
    \caption{\textbf{CLIP-Lite:} Pytorch style pseudo-code for our pretraining framework.}%, reducing the batch size to $O(n)$.}
    \label{fig:pseudo}
    \vspace{-0.1in}
\end{figure}

\subsection{Architecture and Training Details}
% Subsection
In all experiments, we use a standard ResNet-50~\citep{he2016deep} that takes in a $224\times224$ image and generates $2048$-dimensional features at the pre-logit layer. For textual encoding, we use a transformer~\citep{vaswani2017attention} model initialized using  BERT\textsubscript{base}~\citep{devlin2018bert} and use the output \texttt{[CLS]} token as the text representation. We use the COCO Captions dataset~\citep{chen2015microsoft} which has $118$K images with five captions per image. During training time we apply (1) random cropping, (2) color jittering, (3) random horizontal flips while interchanging the words ‘left’ and ‘right’ in the caption, and (4) normalization using the ImageNet image mean. We use SGD with momentum $0.9$~\citep{sutskever2013importance,polyak1964some} and weight decay $10^{-4}$ wrapped in LookAhead~\citep{zhang2019lookahead} with $\alpha=0.5$, and $5$ steps. We perform distributed training across $8$ GPUs with batch normalization~\citep{ioffe2015batch} per GPU with an overall batch size of $1024$ images for $250$K iterations. We use linear learning rate warmup~\citep{goyal2019scaling} for the first $10$K iterations followed by cosine decay~\citep{loshchilov2016sgdr} to zero. Additionally, we train CLIP~\citep{radford2021learning} on the COCO-dataset using an open-source implementation\footnote{\tiny \url{https://github.com/mlfoundations/open\_clip}} with the originally recommended~\citep{radford2021learning} training schedule that suit smaller datasets, reasonable batch-sizes, and compute resources. Specifically, we train using the Adam Optimizer~\citep{kingma2014adam} with decoupled weight decay regularization~\citep{loshchilov2016sgdr} for all weights except gains or biases. We train with a batch-size of $1024$ and warm-up to an initial learning rate of $10^{-4}$ in $10$K steps and decay to zero with the cosine schedule. We found that the performance slightly improves with longer training therefore we train for $250$K iterations, similar to ours. All other training details and hyper-parameters were kept the same as the original work~\citep{radford2021learning}. Please note that our ResNet-50 based CLIP-COCO model outperforms (+1.2\% Zero-shot Acc. on CIFAR10) publicly available weights\footnote{\tiny  \url{https://github.com/revantteotia/clip-training/blob/main/zero_shot_eval_output/coco_trained_clip_observations.md}}, refer to appendix for further details on CLIP-COCO training.

\subsection{Mutual Information Discriminator}
As described in main paper, our JSD-based lower-bound on mutual information relies on a discriminator function, $T_{\omega}: \mathcal{Y} \times \mathcal{Z} \to \mathbb{R}$, which distinguishes between samples extracted from the joint distribution, $P(Y, Z)$~i.e.~a positive image-caption pair and the product of marginals, $P(Y)P(Z)$~i.e.~a negative image-caption pair. This discriminator function can be modelled as an arbitrary neural network with parameters $\omega$ that can be jointly optimized with the encoders during training~\citep{belghazi2018mutual}. In this work, we use a projection and alignment based architecture similar to the one presented in Deep InfoMax~\citep{hjelm2018learning}. 

Given a pair of input one-dimensional representations, both vectors are first projected using a projection module with two linear layers separated by a ReLU and a linear shortcut. A  dot-product of these projections is then computed to get alignment scores. The projection function maps these representations to an aligned cross-modal latent space. Separate projection functions are used for image and text representations. Positive and negative pairs of image-text representations are passed through the discriminator to get respective scores which are then used to estimate and maximize mutual information using our objective. This architecture, in addition to being simple and computationally inexpensive, also offers alignment of the representations into a common cross-modal latent space which uses cosine similarity as the distance metric.

% Frozen backbone table
\begin{table}[ht]
\renewcommand{\arraystretch}{1.2}
\begin{center}
\setlength\tabcolsep{3pt}
\caption{\textbf{Frozen Backbone Results:} On Pascal VOC07 and Imagenet-1k classification, CLIP-Lite outperforms baseline CLIP when evaluated using linear classifiers trained on top of frozen backbone networks pretrained on the COCO Dataset. CLIP-Lite's performance is competitive with more complex vision-language models. CLIP-Lite also performs better than supervised and self-supervised models trained on COCO images, without captions (ref. supplemental materials for additional results).}
\label{tab:transfer}
% \small
% \vspace{-0.1in}
\resizebox{\columnwidth}{!}{%
\begin{tabular}{lcccc}
    \toprule
    \textbf{Method} & \textbf{\# images} & \textbf{Annotations} & \textbf{VOC07} & \textbf{IN-1k} \\ 
    \midrule
    COCO-Sup.      & 118K                & labels            & 86.2           & 46.4           \\
    \midrule
    MoCo-COCO       & 118K               & self-sup.            & 67.5           & 46.5           \\
    \midrule
    ICMLM           & 118K               & captions             & 87.5           & 47.9           \\
    VirTex          & 118K               & captions             & \textbf{88.7}           & \underline{53.8} \\
    \midrule
    CLIP-COCO       & 118K               & captions             & 74.2           & 33.2 \\ 
    CLIP-Lite       & 118K               & captions             & \underline{88.2}           & \textbf{55.3} \\ 
    \bottomrule
\end{tabular}}
\end{center}
\vspace{-0.2in}
\end{table}

% Subsection
\subsection{Transfer Learning with Frozen Backbone}
In these experiments, we train linear models on frozen visual backbones pretrained using CLIP-Lite and compare with  other pretraining methods on PASCAL VOC~\citep{everingham2010pascal} and ImageNet-1k~\citep{russakovsky2015imagenet} classification problems.

\noindent\textbf{PASCAL VOC linear classification:}
For this experiment, our setup is identical to VirTex~\citep{desai2021virtex}. We train on VOC07 trainval split (9K images, 20 classes) and report mAP on the test split. For classification, we train per-class SVMs on 2048-dimensional global average pooled features extracted from the last layer of our trained visual encoder.  For each class, we train SVMs for cost values $C \in \{0.01, 0.1, 1, 10\}$ and select best $C$ by $3$-fold cross-validation.

\noindent\textbf{Imagenet-1k linear classification:}
For this experiment, our setup is identical to VirTex~\citep{desai2021virtex}. We train on the ILSVRC $2012$ train split and report top-$1$ accuracy on val split. We train a linear classifier (fully connected layer + softmax) on $2048$-dimensional global average pooled features extracted from the last layer of the visual backbone. For training, we use a batch-size of $256$ for $100$ epochs. We use SGD with momentum $0.9$ and weight decay $0$. The learning rate schedule is decayed by $0.1$ after $60$ \& $80$ epochs with an initial LR of $30$.

\noindent\textbf{Results:}
We compare CLIP-Lite to supervised, self-supervised and textually-supervised models in Table~\ref{tab:transfer}. CLIP-Lite significantly outperforms baseline CLIP when trained with the same amount of data on both tasks. 
When compared to other image-caption pretraining methods, CLIP-Lite performs competitively with VirTex~\citep{desai2021virtex} on VOC2007 and outperforms both VirTex~\citep{desai2021virtex} and ICMLM~\citep{sariyildiz2020learning}, which are trained on relatively complex language modeling tasks, on Imagenet classification. In addition, different from them, our method also generates a shared latent space that encodes both image and text modalities and enables cheap computation of cross-modal alignment, which enables additional downstream tasks such as zero-shot retrieval, and zero-shot transfer. It also allows us to find subspaces associated with abstract concepts that are better expressed with language than with visual examples, which allows for applications in bias mitigation through the synthesis of gender-neutral image representations. CLIP-Lite also outperforms a fully-supervised model trained with COCO image labels, showing that it learns a better visual representation from information-dense captions as compared to training with labels alone.  Additional results in the supplement show that CLIP-Lite is comparable or better than image-only SSL learning models trained on ImageNet, even though it is trained on much fewer images, albeit with textual supervision.

% ECCV Style table Data Efficiency
\begin{table}[t]
% \small
% \vspace{-0.1in}
\caption{\textbf{Data Efficiency:} CLIP-Lite is more data efficient than CLIP, as shown in this experiment where we pretrain on $\{25, 50, 75, 100\}\%$ of the COCO Captions dataset and evaluate the models on VOC and ImageNet classification tasks with a frozen backbone. CLIP-Lite trained with just 25\% of COCO already surpasses CLIP trained on the whole dataset. }
\label{tab:effi}
\centering
% \vspace{-0.1in}
\resizebox{\columnwidth}{!}{
\begin{tabular}{lccc}
    \toprule
              & \# \textbf{images} & \textbf{VOC07} & \textbf{IN-1k} \\
    \midrule
    % \band
    CLIP COCO-100\%      & 118K      & 74.2       & 33.2 \\
    \midrule
    CLIP-Lite COCO-25\%  & 29.5K     & 77.7\Rise{3.5}        & 45.1\Rise{11.9}        \\
    CLIP-Lite COCO-50\%  & 59K       & 84.4\Rise{10.2}       & 51.3\Rise{18.1}        \\
    CLIP-Lite COCO-75\%  & 88.5K    & 86.8\Rise{12.6}       & 53.2\Rise{20.0}        \\
    CLIP-Lite COCO-100\% & 118K      & 88.2\Rise{14.0}       & 55.3\Rise{22.1}        \\
    \bottomrule
\end{tabular}}
\vspace{-0.1in}
\end{table}

\noindent\textbf{Data Efficiency:} Due to our information-efficient approach for mutual information maximization, CLIP-Lite should be able to learn effective feature representations without requiring as much pretraining data as CLIP. To evaluate this claim, we train ResNet-50 backbones with our pretraining setup on multiple fractional subsets of the COCO Captions dataset and measure their downstream performance on both VOC and ImageNet classification tasks. As demonstrated in Table~\ref{tab:effi}, CLIP-Lite outperforms the original CLIP training objective on VOC  with 20\% and on Imagenet with just $10\%$ of the data, while obtaining a substantial improvement when both are trained with $100\%$ data. Additionally, when compared with Virtex, CLIP-Lite performs competitively on VOC while being consistently better on Imagenet-1k.

\subsection{Transfer Learning with Backbone Finetuning}
Next, we evaluate the performance of of our visual backbone when the entire network is finetuned for the downstream task. For this purpose, we perform fine-grained classification on the iNaturalist 2018~\citep{van2018inaturalist} dataset, which contains images from $8,142$ fine-grained categories, with a long-tailed distribution. We train with the `train2018' split and evaluate in the `val2018' split. We finetune pretrained ResNet-$50$ models with a linear layer, using SGD with momentum $0.9$ and weight decay $10^{-4}$ for $100$ epochs. Initial learning rate is set to $0.025$, which is reduced by $10\times$ at epochs $70$ and $90$. We use a batch size of $256$ distributed across $8$ GPUs.

\begin{table}[t]
% \vspace{-0.1in}
\caption{\textbf{Backbone Finetuning Results:} CLIP-Lite outperforms CLIP-COCO on iNaturalist, and performs comparably to VirTex. (IN-Sup. = ImageNet-supervised.)}
\label{tab:inat}
\centering
% \vspace{-0.1in}
\resizebox{\columnwidth}{!}{
    \begin{tabular}{lccc}
        \toprule
        \textbf{Method} & \textbf{\# images} & \textbf{Annotations} & \textbf{iNat 18} \\
        \midrule
        Random Init     & -                  & -                    & 61.4             \\
        \midrule
        IN-sup          & 1.28M              & labels               & 65.2             \\
        IN-sup-50\%     & 640K               & labels               & 63.2             \\
        IN-sup-10\%     & 128K               & labels               & 60.2             \\
        \midrule
        MoCo-COCO       & 118K               & self-sup.            & 60.5             \\
        MoCo-IN         & 1.28M              & self-sup.            & 63.2             \\
        \midrule
        VirTex          & 118K               & captions             & 63.4             \\\midrule
        CLIP-COCO       & 118K               & captions             & 61.8             \\
        
        CLIP-Lite       & 118K               & captions             & 63.1            \\
        \bottomrule
    \end{tabular}
}
\vspace{-0.22in}
\end{table}

\noindent\textbf{Results:}
We summarize our results in Table~\ref{tab:inat}. CLIP-Lite is competitive with supervised and self-supervised learning models trained with images alone even those trained with 5-10x more images. Its performance matches closely a model trained with full-supervision on $50\%$ of the ImageNet~\citep{krizhevsky2012imagenet} dataset, equal to $5.4\times$ the number of images as our pretraining dataset. Finally, CLIP-Lite obtains a $1.3\%$ improvement over CLIP-COCO, while being competitive with VirTex.

% ECCV Style table
\begin{table*}[ht]
\vspace{-0.1in}
\caption{\textbf{Retrieval Results:} CLIP-Lite substantially outperforms CLIP-COCO and the baseline Visual N-grams~\citep{li2017learning} approach. CLIP-Lite is superior when evaluated on  the COCO test split, which is similar to the CLIP-Lite training set and on Flickr30K, generalizing to unseen images and text in a zero-shot manner.}
% \vspace{-0.1in}
\label{tab:retrieval}
\centering
    \begin{tabularx}{\textwidth}{lcccccccccccc}
    \toprule
                    & \multicolumn{6}{c}{\textbf{Text Retrieval}}                                   & \multicolumn{6}{c}{\textbf{Image Retrieval}}                                  \\ \cmidrule{2-13} 
                    & \multicolumn{3}{c}{\textbf{Flickr30k}} & \multicolumn{3}{c}{\textbf{MSCOCO}} & \multicolumn{3}{c}{\textbf{Flickr30k}} & \multicolumn{3}{c}{\textbf{MSCOCO}} \\ \cmidrule{2-13}
    
    \textbf{Method} & {\it R@1} & {\it R@5} & {\it R@10} & 
                    {\it R@1} & {\it R@5} & {\it R@10} & 
                    {\it R@1} & {\it R@5} & {\it R@10} & 
                    {\it R@1} & {\it R@5} & {\it R@10} \\ 
                    \midrule
    Visual N-Grams  & 15.4 & 35.7 & 45.1 & 8.7 & 23.1 & 33.3 & 8.8 & 21.2 & 29.9 & 5.0 & 14.5 & 21.9 \\
    CLIP-COCO       & 19.9 & 41.9 & 54.9 & 18.9 & 42.9 & 54.6 & 13.9 & 33.0 & 43.8 & 13.9 & 33.5 & 44.2 \\ \midrule
    CLIP-Lite       & \textbf{28.8} & \textbf{55.8} & \textbf{67.4} & \textbf{26.0} & \textbf{54.6} & \textbf{68.0} & \textbf{23.1} & \textbf{51.1} & \textbf{62.9} & \textbf{20.2} & \textbf{48.1} & \textbf{62.2} \\ 
    \bottomrule
\end{tabularx}
\vspace{-0.1in}
\end{table*}

% Subsection
\subsection{Image-Text and Text-Image Retrieval} \label{sec:retrieval}
Our method is expected to produce effective representations for the task of image-text retrieval as it is trained by aligning text and image representations. We evaluate the image-text retrieval capabilities of CLIP-Lite on the validation set of COCO and the test split of Flickr30k~\citep{young2014image} datasets, following CLIP. We perform zero-shot image-text and text-image retrieval by ranking image-text pairs by their alignment score, which is the dot product of the normalized representations in the shared latent space. This ability to perform zero-shot retrieval is a salient feature of our and CLIP-like methods over previously proposed works that rely on language modeling tasks.

\noindent\textbf{Results:} Table~\ref{tab:retrieval} shows that CLIP-Lite substantially outperforms CLIP-COCO on all metrics for both text and image retrieval. The performance improvement is large both when evaluated on the COCO validation set, which is similar to the the COCO-Captions training split used for CLIP-Lite training; and when testing zero-shot on unseen text vocabulary and object categories of Flickr30K. Taken together, these results show that CLIP-Lite learns a superior representation for retrieval tasks as compared to CLIP, when trained on same amounts of data.

\subsection{Zero-Shot Transfer}
We use the cross-modal alignment capability of CLIP-Lite to perform zero-shot classification on unseen datasets CIFAR-10, CIFAR100~\citep{krizhevsky2009learning}, ImageNetV2~\citep{recht2019imagenet}, and ImageNet-A~\citep{hendrycks2021natural}. Our model generates a shared latent space where we can readily compute the alignment between given (image, text) pairs as the cosine similarity of their representations. Therefore, we use the names of the classes to generate a textual description of each class label (class prompt). In this experiment, we use templates such as, ``a photo of a \{class name\}'' to generate such class prompts, following CLIP~\citep{radford2021learning}. Please refer to the appendix for comparison between different templates for generating the prompts. For a given image, we compute its alignment with each of the class prompts which are then normalized into a probability distribution via a softmax.

\noindent\textbf{Results:} Our results for the zero-shot transfer task on unseen datasets are compiled in table~\ref{tab:zero}. Given the zero-shot nature of the task, CLIP-Lite obtains satisfactory performance on the complex ImageNet evaluations while clearly outperforming CLIP trained with the same amount of data in all settings.

\subsection{Evaluating Visual Grounding}
% \vspace{-0.0in}
Next, we evaluate the capability of CLIP-Lite to localize a region in the image that corresponds to a given textual description. We compute the dot-product of the visual and textual embedding and compute its gradients with respect to the last convolutional layer of ResNet. We global average pool these gradients and perform a weighted sum with the last convolutional activations and clip the negative values to obtain Grad-CAM~\citep{selvaraju2017grad}. We then use the areas highlighted by Grad-CAM to approximate a predicted bounding box. We evaluate this experiment on the RefCOCO+~\citep{yu2016modeling} dataset. We note that the images in the RefCOCO+ dataset are extracted from the training set of the COCO~\citep{chen2015microsoft} dataset which our model uses for pretraining. Therefore, we view this evaluation as an explorative study to establish that our model is focusing on the relevant areas of the image while computing the alignment score with the caption.

\noindent
RefCOCO+ results can be seen in the table to the right. CLIP-Lite significantly outperforms CLIP on all settings. 

\begin{wraptable}{r}{0.6\columnwidth}
% \begin{table}[!t]
\small
% \vspace{-0.1in}
% \centering	
\setlength\tabcolsep{1pt}
\renewcommand{\arraystretch}{1.0}
     \resizebox{0.6\columnwidth}{!}{%
		\begin{tabular}	{l   c c c }
			\toprule
			Method & Val-acc & TestA-acc & TestB-acc \\
			\midrule
			CLIP-COCO & 29.1 & 28.5 & 28.5 \\
			CLIP-Lite (ours) & \textbf{36.1} & \textbf{41.4} & \textbf{32.0}\\			
			\bottomrule
			% \vspace{-0.1in}
		\end{tabular}
		}
\end{wraptable}

\begin{figure*}[t]
\vspace{-0.12in}
\begin{center}
\includegraphics[width=0.99\linewidth]{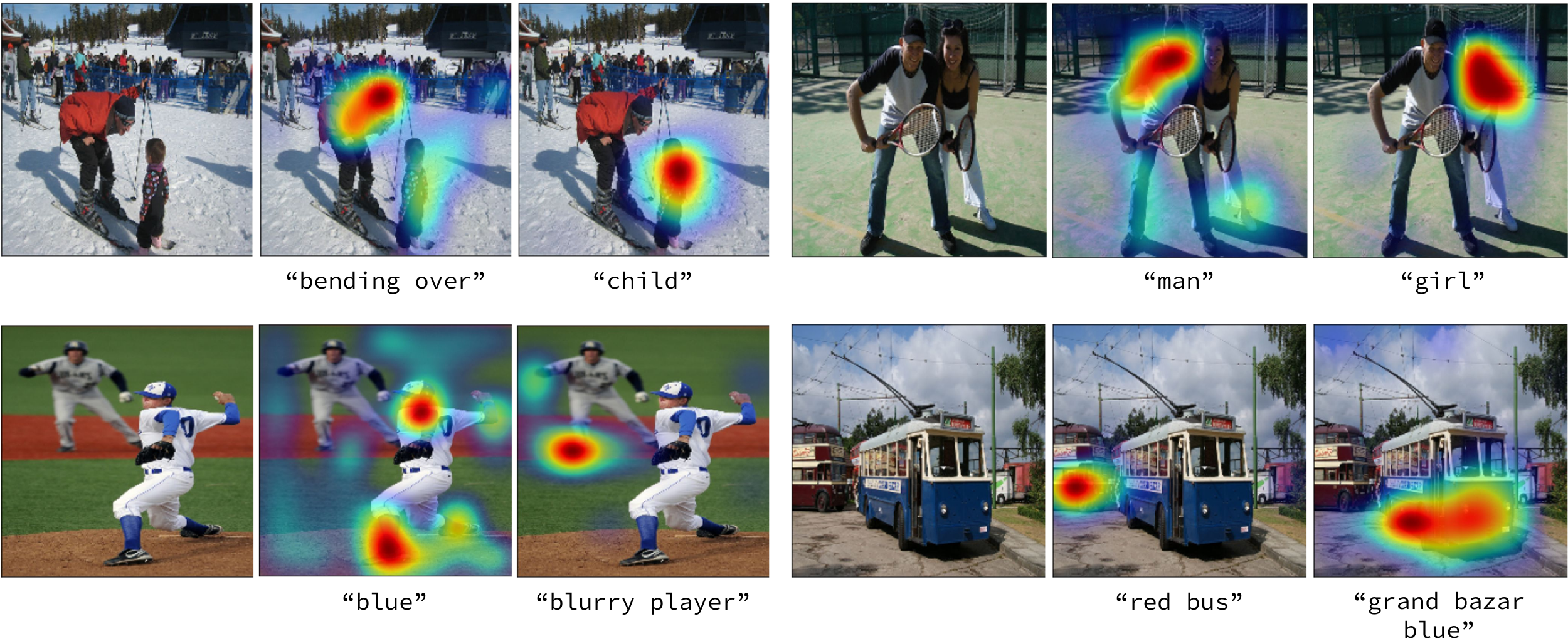}
% \vspace{-0.2in}
\end{center}
   \caption{\textbf{Visual Grounding on RefCOCO+:} CLIP-Lite is able to localize textual descriptions to relevant areas in the image, shown here through Grad-CAM visualization using the alignment score with the mentioned textual description. \textit{Top left:} CLIP-Lite  is able to localize the action phrases such as ``bending over''. This demonstrates the value of learning from semantically rich textual captions. }
\label{fig:refcoco}
\vspace{-0.1in}
\end{figure*}

Qualitative results in Figure~\ref{fig:refcoco} demonstrate that even though the network has not been trained with any localization supervision, it is surprisingly good at localizing phrases in the image. For instance, in Figure~\ref{fig:refcoco} bottom left, for the phrase ``blue", the network attends to all blue regions in the player's outfit. Interestingly, it is also able to localize abstract concepts as ``blurry player". 

\begin{table}[t]
\renewcommand{\arraystretch}{1.1}
\begin{center}
\tiny
\caption{\textbf{Zero Shot Transfer:} CLIP-Lite obtains satisfactory zero-shot transfer to unseen datasets.}
% \vspace{-0.1in}
\label{tab:zero}
\resizebox{\columnwidth}{!}{%
    \begin{tabular}{lcccc}
    \toprule
                    & \multicolumn{2}{c}{\textbf{CLIP-COCO}} 
                    & \multicolumn{2}{c}{\textbf{CLIP-Lite}} \\
                    \midrule
                    % \cmidrule{2-13}
    
    \textbf{Dataset} & {\it Top1} & {\it Top5} & 
                    {\it Top1} & {\it Top5} \\ 
                    \midrule
                    
    CIFAR10  & 16.3 & 68.9 & \textbf{33.0} & \textbf{82.7} \\
    CIFAR100 & 2.9 & 12.4 & \textbf{6.8} & \textbf{33.1} \\
    ImageNet-V2 & 4.4 & 11.1 & \textbf{9.9} & \textbf{21.4} \\
    ImageNet-A & 1.7 & 7.3 & \textbf{3.8} & \textbf{14.9} \\
    \bottomrule
\end{tabular}
}
\end{center}
% \vspace{-0.3in}
\end{table}

% Subsection
\subsection{Editing Concepts from Image Representations}
One salient feature of CLIP-like methods, which other methods such as VirTex~\citep{desai2021virtex} and ICMLM~\citep{sariyildiz2020learning} lack, is that they are able to generate a shared latent space that encodes both image and text modalities. %In addition, it is able to cheaply compute cross-modal alignment. 
This enables us to find representations and subspaces associated with abstract concepts that are better expressed with language than with visual examples. Using this property, we demonstrate a methodology to remove concepts from visual representations. For instance, it is non trivial and even problematic to collect visual examples that capture the concept of gender, while it is relatively straightforward to express this concept in a sentence using language. Therefore, we can identify the gender subspace in our shared embedding space using text and use it to remove variance along this direction to smooth out the concept of gender from image representations. We motivate this experiment in the growing body of literature regarding bias mitigation, where the objective is to build invariant representations with respect to sensitive or protected attributes~\citep{wang2019balanced,wang2020towards}. In comparison to our work other methods require retraining the models to obtain invariant bias representations through adversarial learning~\citep{wang2019balanced} or effectively combining domain independent classifiers~\citep{wang2020towards}. 

\begin{figure*}[t]
\vspace{-0.15in}
\begin{center}
\includegraphics[width=0.99\linewidth]{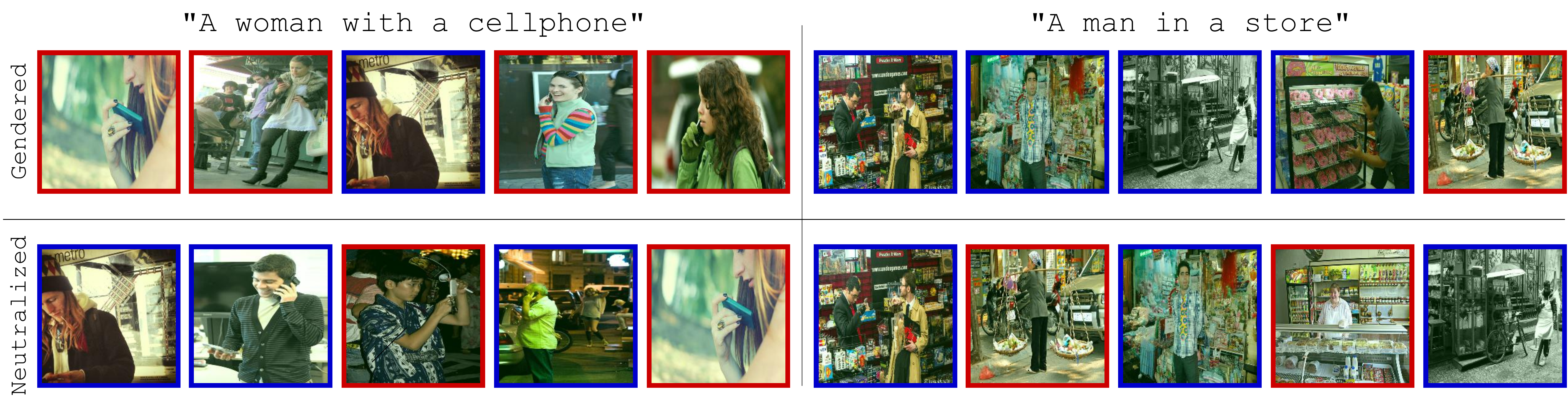}
\end{center}
   \caption{\textbf{Demonstrating Neutral Representations:} Qualitative demonstration of our concept editing method. For each text prompt, most aligned images are retrieved from male and female buckets of the gendered COCO subset before (top row) and after (bottom row) gender smoothing. Once representations are gender-neutralized the gendered references in the query become irrelevant and the image is only retrieved based on its remaining contents. Alignment score decreases from left to right for each set of queried images.
   Boundary color denotes perceived image gender; red for female, blue for male.}
\label{fig:bias}
\vspace{-0.2in}
\end{figure*}

\noindent\textbf{Identifying the Concept Subspace:}
The first step of our approach is to isolate the direction in the embedding space that captures maximum gender variance. For this purpose, we follow a strategy similar to Bolukbasi et al.~\citep{bolukbasi2016man} that deals with debiasing word representations. For characterizing features for male and female genders, we use word pairs (\textit{man, woman}), (\textit{son, daughter}) that indicate opposite genders. Now, consider a dataset $\mathcal{D} = \{(w_m, w_f)\}_{i=1}^{m}$ where each entry $(w_m, w_f)$ is a tuple of opposite gendered words. Intuitively, each tuple should contain words that have the same meaning if not for the target attribute. To make the set $\mathcal{D}$ more robust, we used the sentence contextualization strategy presented in Liang et al.~\citep{liang2020towards}. In this step, the predefined sets of gendered tokens in the set, $\mathcal{D}$, are used to generate paired sentences which have the same meaning except for the gender attribute. We perform this contextualization by using simple sentence templates such as ``I am a [word]'' where [word] can be replaced with the word pairs in our dataset $\mathcal{D}$ to give, for instance, (\textit{``I am a boy.'', ``I am a girl.''}). Hence, we obtain a contextualized bias attribute dataset $\mathcal{S} = \{(s_m, s_f)\}_{i=1}^{n}$ where each entry is a tuple of semantically similar sentences with opposite genders. We extract the sentence representations for all entries in the set $\mathcal{S}$ by passing them through our pretrained text encoder and then projecting them to the shared latent space using the projector trained with our mutual information discriminator $T_\omega$. We define sets $\mathcal{R}_m$ and $\mathcal{R}_f$ that contain sentence representations of the male and the female category, for example, $\mathcal{R}_m = \{ F_t(s_m) \}_{i=1}^n$ where $F_t(.)$ is the sequential combination of our pretrained text-encoder and text-projection functions. Now we estimate the gender subspace $V = \{v_1, ..., v_k \}$ using the Principal Component Analysis corresponding mean shifted representation from both sets as described in~\citep{liang2020towards}.

\begin{table}[t]
%\vspace{-20pt}
% \begin{table}[!t]
\caption{\textbf{Concept Editing Results:} We compute the mean alignment scores for the top 10 images queried using prompts that either contain male or female gendered tokens. The images are queried using gendered and neutralized representations. We observe that after gender-deletion the alignment score for images with men and women converge to similar values.}
% \vspace{-0.1in}
\label{tab:debiasing}
\small
\centering	
\setlength\tabcolsep{1.5pt}
\renewcommand{\arraystretch}{1.2}
     \resizebox{\columnwidth}{!}{%
		\begin{tabular}{lcccccccc}
        \toprule
            && \multicolumn{3}{c}{\textbf{Images with Men}} && \multicolumn{3}{c}{\textbf{Images with Women}} \\
        \cmidrule{3-5}\cmidrule{7-9}
           && gendered       & neutral       & delta        && gendered     & neutral     & delta     \\
        \midrule
        Male queries           && 0.085        & 0.069          & +0.016       && 0.057               & 0.067        & -0.010    \\
        Female queries         && 0.042        & 0.068          & -0.026       && 0.089               & 0.062        & +0.027 \\
        \bottomrule
        \end{tabular}
		}
\vspace{-0.3in}
\end{table}

\noindent\textbf{Removing Concept from Image Representations:}
After estimating the gender subspace in our shared cross-modal latent space, we extend the hard debias algorithm~\citep{bolukbasi2016man} to edit visual representations. This is achieved by first projecting the representation onto the bias subspace, this projection is then subtracted from the original representation to give the de-gendered representation. Given an image, we first encode the image onto our multi-modal shared latent space to get, say, $h$. Now, consider the identified gender subspace $V$, we first compute the projection of $h$ onto this gender subspace $V$ to get $h_{V} = \sum_{j=1}^{k} \left \langle h, v_{j} \right \rangle v_j$. We subtract this projection from the original representation to get a vector, $\hat{h} = h - h_{V}$ that is orthogonal to the bias subspace and therefore does not encode the target bias.

\noindent\textbf{Analysis:}
To evaluate concept editing, we use the gendered subset of  COCO-Captions~\citep{wang2019balanced,zhao2017men} for studying bias. The gender labels for images in the COCO dataset are derived from the captions. We obtain a subset from the COCO dataset with $16,225$ images with men and $6,601$ images with women. We use $10$ sentences with male references and $10$ sentences with female references from the set $\mathcal{S}$ and use them as prompts for this study. For each gendered prompt, we query the top $10$ images independently from the male and the female image sets using both biased and debiased representations to compute alignment with the prompt. The mean alignment scores are then computed for each set given the prompt. Table~\ref{tab:debiasing} shows that the alignment scores roughly equalize for members of the two groups after removing the variance along the gender direction from the visual representations which indicates the invariance of the visual representations to gendered language tokens. 
\vspace{-0.1in}

\section{Limitations and Broader Impacts}
\vspace{-0.05in}
CLIP-Lite trains the visual encoder by maximizing the mutual information between images and their captions. We observe that language supervision provides rich semantic density which can be distilled into visual representations. The visual encoder is encouraged to learn visual representations that encode maximum information from captions. As such, the visual encoder is only aware of concepts and objects that human-annotators have mentioned in the captions. Therefore the visual encoder lags behind task-specific models that are trained specifically for a given fine-grained task. For instance, visual encoders trained with CLIP-Lite struggle with relatively contextual downstream tasks that involve reading text or counting number of objects in an image. In this work, we train CLIP-Lite on the COCO-Captions~\citep{chen2015microsoft} dataset which has high-quality curated captions for images. However, when trained on datasets with text paired with images from the internet, the textual captions can be significantly unfiltered and noisy. Our method essentially learns by aligning the caption and text representations. Therefore, the model is susceptible to learning harmful biases that are represented in the captions. Hence, deployment of visual backbones trained  with CLIP-Lite and other pretraining methods which use natural language supervision need to be analyzed specifically for such biases. In this work, we present an approach to edit concepts from visual representations using the shared vision-language latent space learnt by our method. For instance, we demonstrate this capability by editing visual representations such that they are invariant to gendered tokens in language. However, further explorations are required to develop this concept editing mechanism further.

% \vspace{-0.02in}
\section{Conclusion}
\label{sec:discussion}
\vspace{-0.05in}
We introduced CLIP-Lite an image-text pretrained model using contrastive learning that leverages a different objective than the CLIP model that allows for it to be more data efficient. CLIP-Lite's objective is insensitive to the number of negative samples and hence can be trained with just one negative image-caption pair and shows superior results on lower data regimes while still demonstrating some of the most remarkable capabilities of the original CLIP model such as transferable features, zero-shot capabilities, and a shared latent space. Additionally, we present a concept editing methodology for neutralizing visual representations with respect to a chosen abstract concept. Please refer to the supplement for a detailed discussion on limitations and potential impact of our approach.

\textbf{Acknowledgments} 
This work was supported by NSF Awards IIS-2221943 and IIS-2201710, and through gift funding from a Salesforce AI Research Grant.

%%%%%%%%% APPENDIX
\clearpage
\bibliography{egbib}

\begin{thebibliography}{}

\bibitem[Antol et~al., 2015]{antol2015vqa}
Antol, S., Agrawal, A., Lu, J., Mitchell, M., Batra, D., Zitnick, C.~L., and
  Parikh, D. (2015).
\newblock Vqa: Visual question answering.
\newblock In {\em Proceedings of the IEEE international conference on computer
  vision}, pages 2425--2433.

\bibitem[Bachman et~al., 2019]{bachman2019learning}
Bachman, P., Hjelm, R.~D., and Buchwalter, W. (2019).
\newblock Learning representations by maximizing mutual information across
  views.
\newblock {\em arXiv preprint arXiv:1906.00910}.

\bibitem[Belghazi et~al., 2018]{belghazi2018mutual}
Belghazi, M.~I., Baratin, A., Rajeshwar, S., Ozair, S., Bengio, Y., Courville,
  A., and Hjelm, D. (2018).
\newblock Mutual information neural estimation.
\newblock In {\em International Conference on Machine Learning}, pages
  531--540. PMLR.

\bibitem[Bolukbasi et~al., 2016]{bolukbasi2016man}
Bolukbasi, T., Chang, K.-W., Zou, J.~Y., Saligrama, V., and Kalai, A.~T.
  (2016).
\newblock Man is to computer programmer as woman is to homemaker? debiasing
  word embeddings.
\newblock {\em Advances in neural information processing systems},
  29:4349--4357.

\bibitem[Brakel and Bengio, 2017]{brakel2017learning}
Brakel, P. and Bengio, Y. (2017).
\newblock Learning independent features with adversarial nets for non-linear
  ica.
\newblock {\em arXiv preprint arXiv:1710.05050}.

\bibitem[Caron et~al., 2020]{caron2020unsupervised}
Caron, M., Misra, I., Mairal, J., Goyal, P., Bojanowski, P., and Joulin, A.
  (2020).
\newblock Unsupervised learning of visual features by contrasting cluster
  assignments.
\newblock {\em arXiv preprint arXiv:2006.09882}.

\bibitem[Chen et~al., 2020a]{chen2020simple}
Chen, T., Kornblith, S., Norouzi, M., and Hinton, G. (2020a).
\newblock A simple framework for contrastive learning of visual
  representations.
\newblock In {\em International conference on machine learning}, pages
  1597--1607. PMLR.

\bibitem[Chen et~al., 2020b]{chen2020improved}
Chen, X., Fan, H., Girshick, R., and He, K. (2020b).
\newblock Improved baselines with momentum contrastive learning.
\newblock {\em arXiv preprint arXiv:2003.04297}.

\bibitem[Chen et~al., 2015]{chen2015microsoft}
Chen, X., Fang, H., Lin, T.-Y., Vedantam, R., Gupta, S., Doll{\'a}r, P., and
  Zitnick, C.~L. (2015).
\newblock Microsoft coco captions: Data collection and evaluation server.
\newblock {\em arXiv preprint arXiv:1504.00325}.

\bibitem[Desai and Johnson, 2021]{desai2021virtex}
Desai, K. and Johnson, J. (2021).
\newblock Virtex: Learning visual representations from textual annotations.
\newblock In {\em Proceedings of the IEEE/CVF Conference on Computer Vision and
  Pattern Recognition}, pages 11162--11173.

\bibitem[Devlin et~al., 2018]{devlin2018bert}
Devlin, J., Chang, M.-W., Lee, K., and Toutanova, K. (2018).
\newblock Bert: Pre-training of deep bidirectional transformers for language
  understanding.
\newblock {\em arXiv preprint arXiv:1810.04805}.

\bibitem[Donsker and Varadhan, 1983]{donsker1983asymptotic}
Donsker, M.~D. and Varadhan, S.~S. (1983).
\newblock Asymptotic evaluation of certain markov process expectations for
  large time. iv.
\newblock {\em Communications on Pure and Applied Mathematics}, 36(2):183--212.

\bibitem[Everingham et~al., 2010]{everingham2010pascal}
Everingham, M., Van~Gool, L., Williams, C.~K., Winn, J., and Zisserman, A.
  (2010).
\newblock The pascal visual object classes (voc) challenge.
\newblock {\em International journal of computer vision}, 88(2):303--338.

\bibitem[Girshick et~al., 2014]{girshick2014rich}
Girshick, R., Donahue, J., Darrell, T., and Malik, J. (2014).
\newblock Rich feature hierarchies for accurate object detection and semantic
  segmentation.
\newblock In {\em Proceedings of the IEEE conference on computer vision and
  pattern recognition}, pages 580--587.

\bibitem[Goyal et~al., 2019]{goyal2019scaling}
Goyal, P., Mahajan, D., Gupta, A., and Misra, I. (2019).
\newblock Scaling and benchmarking self-supervised visual representation
  learning.
\newblock In {\em Proceedings of the IEEE/CVF International Conference on
  Computer Vision}, pages 6391--6400.

\bibitem[Grill et~al., 2020]{grill2020bootstrap}
Grill, J.-B., Strub, F., Altch{\'e}, F., Tallec, C., Richemond, P.~H.,
  Buchatskaya, E., Doersch, C., Pires, B.~A., Guo, Z.~D., Azar, M.~G., et~al.
  (2020).
\newblock Bootstrap your own latent: A new approach to self-supervised
  learning.
\newblock {\em arXiv preprint arXiv:2006.07733}.

\bibitem[Gutmann and Hyv{\"a}rinen, 2010]{gutmann2010noise}
Gutmann, M. and Hyv{\"a}rinen, A. (2010).
\newblock Noise-contrastive estimation: A new estimation principle for
  unnormalized statistical models.
\newblock In {\em Proceedings of the Thirteenth International Conference on
  Artificial Intelligence and Statistics}, pages 297--304. JMLR Workshop and
  Conference Proceedings.

\bibitem[He et~al., 2020]{he2020momentum}
He, K., Fan, H., Wu, Y., Xie, S., and Girshick, R. (2020).
\newblock Momentum contrast for unsupervised visual representation learning.
\newblock In {\em Proceedings of the IEEE/CVF Conference on Computer Vision and
  Pattern Recognition}, pages 9729--9738.

\bibitem[He et~al., 2016]{he2016deep}
He, K., Zhang, X., Ren, S., and Sun, J. (2016).
\newblock Deep residual learning for image recognition.
\newblock In {\em Proceedings of the IEEE conference on computer vision and
  pattern recognition}, pages 770--778.

\bibitem[Hendrycks et~al., 2021]{hendrycks2021natural}
Hendrycks, D., Zhao, K., Basart, S., Steinhardt, J., and Song, D. (2021).
\newblock Natural adversarial examples.
\newblock In {\em Proceedings of the IEEE/CVF Conference on Computer Vision and
  Pattern Recognition}, pages 15262--15271.

\bibitem[Hjelm et~al., 2018]{hjelm2018learning}
Hjelm, R.~D., Fedorov, A., Lavoie-Marchildon, S., Grewal, K., Bachman, P.,
  Trischler, A., and Bengio, Y. (2018).
\newblock Learning deep representations by mutual information estimation and
  maximization.
\newblock {\em arXiv preprint arXiv:1808.06670}.

\bibitem[Ioffe and Szegedy, 2015]{ioffe2015batch}
Ioffe, S. and Szegedy, C. (2015).
\newblock Batch normalization: Accelerating deep network training by reducing
  internal covariate shift.
\newblock In {\em International conference on machine learning}, pages
  448--456. PMLR.

\bibitem[Jabri et~al., 2020]{jabri2020space}
Jabri, A., Owens, A., and Efros, A.~A. (2020).
\newblock Space-time correspondence as a contrastive random walk.
\newblock {\em arXiv preprint arXiv:2006.14613}.

\bibitem[Joulin et~al., 2016]{joulin2016learning}
Joulin, A., Van Der~Maaten, L., Jabri, A., and Vasilache, N. (2016).
\newblock Learning visual features from large weakly supervised data.
\newblock In {\em European Conference on Computer Vision}, pages 67--84.
  Springer.

\bibitem[Kingma and Ba, 2014]{kingma2014adam}
Kingma, D.~P. and Ba, J. (2014).
\newblock Adam: A method for stochastic optimization.
\newblock {\em arXiv preprint arXiv:1412.6980}.

\bibitem[Krizhevsky et~al., 2009]{krizhevsky2009learning}
Krizhevsky, A., Hinton, G., et~al. (2009).
\newblock Learning multiple layers of features from tiny images.

\bibitem[Krizhevsky et~al., 2012]{krizhevsky2012imagenet}
Krizhevsky, A., Sutskever, I., and Hinton, G.~E. (2012).
\newblock Imagenet classification with deep convolutional neural networks.
\newblock {\em Advances in neural information processing systems},
  25:1097--1105.

\bibitem[Lei~Ba et~al., 2015]{lei2015predicting}
Lei~Ba, J., Swersky, K., Fidler, S., et~al. (2015).
\newblock Predicting deep zero-shot convolutional neural networks using textual
  descriptions.
\newblock In {\em Proceedings of the IEEE International Conference on Computer
  Vision}, pages 4247--4255.

\bibitem[Li et~al., 2017]{li2017learning}
Li, A., Jabri, A., Joulin, A., and van~der Maaten, L. (2017).
\newblock Learning visual n-grams from web data.
\newblock {\em Proceedings of the IEEE International Conference on Computer
  Vision}, pages 4183--4192.

\bibitem[Liang et~al., 2020]{liang2020towards}
Liang, P.~P., Li, I.~M., Zheng, E., Lim, Y.~C., Salakhutdinov, R., and Morency,
  L.-P. (2020).
\newblock Towards debiasing sentence representations.
\newblock {\em arXiv preprint arXiv:2007.08100}.

\bibitem[Long et~al., 2015]{long2015fully}
Long, J., Shelhamer, E., and Darrell, T. (2015).
\newblock Fully convolutional networks for semantic segmentation.
\newblock In {\em Proceedings of the IEEE conference on computer vision and
  pattern recognition}, pages 3431--3440.

\bibitem[Loshchilov and Hutter, 2016]{loshchilov2016sgdr}
Loshchilov, I. and Hutter, F. (2016).
\newblock Sgdr: Stochastic gradient descent with warm restarts.
\newblock {\em arXiv preprint arXiv:1608.03983}.

\bibitem[McAllester and Stratos, 2020]{mcallester2020formal}
McAllester, D. and Stratos, K. (2020).
\newblock Formal limitations on the measurement of mutual information.
\newblock In {\em International Conference on Artificial Intelligence and
  Statistics}, pages 875--884. PMLR.

\bibitem[Mikolov et~al., 2013]{mikolov2013efficient}
Mikolov, T., Chen, K., Corrado, G., and Dean, J. (2013).
\newblock Efficient estimation of word representations in vector space.
\newblock {\em arXiv preprint arXiv:1301.3781}.

\bibitem[Nowozin et~al., 2016]{nowozin2016f}
Nowozin, S., Cseke, B., and Tomioka, R. (2016).
\newblock f-gan: Training generative neural samplers using variational
  divergence minimization.
\newblock In {\em Proceedings of the 30th International Conference on Neural
  Information Processing Systems}, pages 271--279.

\bibitem[Oord et~al., 2018]{oord2018representation}
Oord, A. v.~d., Li, Y., and Vinyals, O. (2018).
\newblock Representation learning with contrastive predictive coding.
\newblock {\em arXiv preprint arXiv:1807.03748}.

\bibitem[Polyak, 1964]{polyak1964some}
Polyak, B.~T. (1964).
\newblock Some methods of speeding up the convergence of iteration methods.
\newblock {\em Ussr computational mathematics and mathematical physics},
  4(5):1--17.

\bibitem[Quattoni et~al., 2007]{quattoni2007learning}
Quattoni, A., Collins, M., and Darrell, T. (2007).
\newblock Learning visual representations using images with captions.
\newblock In {\em 2007 IEEE Conference on Computer Vision and Pattern
  Recognition}, pages 1--8. IEEE.

\bibitem[Radford et~al., 2021]{radford2021learning}
Radford, A., Kim, J.~W., Hallacy, C., Ramesh, A., Goh, G., Agarwal, S., Sastry,
  G., Askell, A., Mishkin, P., Clark, J., et~al. (2021).
\newblock Learning transferable visual models from natural language
  supervision.
\newblock {\em arXiv preprint arXiv:2103.00020}.

\bibitem[Recht et~al., 2019]{recht2019imagenet}
Recht, B., Roelofs, R., Schmidt, L., and Shankar, V. (2019).
\newblock Do imagenet classifiers generalize to imagenet?
\newblock In {\em International Conference on Machine Learning}, pages
  5389--5400. PMLR.

\bibitem[Russakovsky et~al., 2015]{russakovsky2015imagenet}
Russakovsky, O., Deng, J., Su, H., Krause, J., Satheesh, S., Ma, S., Huang, Z.,
  Karpathy, A., Khosla, A., Bernstein, M., et~al. (2015).
\newblock Imagenet large scale visual recognition challenge.
\newblock {\em International journal of computer vision}, 115(3):211--252.

\bibitem[Sariyildiz et~al., 2020]{sariyildiz2020learning}
Sariyildiz, M.~B., Perez, J., and Larlus, D. (2020).
\newblock Learning visual representations with caption annotations.
\newblock In {\em Computer Vision--ECCV 2020: 16th European Conference,
  Glasgow, UK, August 23--28, 2020, Proceedings, Part VIII 16}, pages 153--170.
  Springer.

\bibitem[Selvaraju et~al., 2017]{selvaraju2017grad}
Selvaraju, R.~R., Cogswell, M., Das, A., Vedantam, R., Parikh, D., and Batra,
  D. (2017).
\newblock Grad-cam: Visual explanations from deep networks via gradient-based
  localization.
\newblock In {\em Proceedings of the IEEE international conference on computer
  vision}, pages 618--626.

\bibitem[Stratos, 2018]{stratos2018mutual}
Stratos, K. (2018).
\newblock Mutual information maximization for simple and accurate
  part-of-speech induction.
\newblock {\em arXiv preprint arXiv:1804.07849}.

\bibitem[Stroud et~al., 2020]{stroud2020learning}
Stroud, J.~C., Lu, Z., Sun, C., Deng, J., Sukthankar, R., Schmid, C., and Ross,
  D.~A. (2020).
\newblock Learning video representations from textual web supervision.
\newblock {\em arXiv preprint arXiv:2007.14937}.

\bibitem[Sutskever et~al., 2013]{sutskever2013importance}
Sutskever, I., Martens, J., Dahl, G., and Hinton, G. (2013).
\newblock On the importance of initialization and momentum in deep learning.
\newblock In {\em International conference on machine learning}, pages
  1139--1147. PMLR.

\bibitem[Tian et~al., 2019]{tian2019contrastive}
Tian, Y., Krishnan, D., and Isola, P. (2019).
\newblock Contrastive representation distillation.
\newblock {\em arXiv preprint arXiv:1910.10699}.

\bibitem[Tian et~al., 2020]{tian2020makes}
Tian, Y., Sun, C., Poole, B., Krishnan, D., Schmid, C., and Isola, P. (2020).
\newblock What makes for good views for contrastive learning?
\newblock {\em arXiv preprint arXiv:2005.10243}.

\bibitem[Tishby and Zaslavsky, 2015]{tishby2015deep}
Tishby, N. and Zaslavsky, N. (2015).
\newblock Deep learning and the information bottleneck principle.
\newblock In {\em 2015 IEEE Information Theory Workshop (ITW)}, pages 1--5.
  IEEE.

\bibitem[Van~Horn et~al., 2018]{van2018inaturalist}
Van~Horn, G., Mac~Aodha, O., Song, Y., Cui, Y., Sun, C., Shepard, A., Adam, H.,
  Perona, P., and Belongie, S. (2018).
\newblock The inaturalist species classification and detection dataset.
\newblock In {\em Proceedings of the IEEE conference on computer vision and
  pattern recognition}, pages 8769--8778.

\bibitem[Vaswani et~al., 2017]{vaswani2017attention}
Vaswani, A., Shazeer, N., Parmar, N., Uszkoreit, J., Jones, L., Gomez, A.~N.,
  Kaiser, {\L}., and Polosukhin, I. (2017).
\newblock Attention is all you need.
\newblock In {\em Advances in neural information processing systems}, pages
  5998--6008.

\bibitem[Veli{\v{c}}kovi{\'c} et~al., 2018]{velivckovic2018deep}
Veli{\v{c}}kovi{\'c}, P., Fedus, W., Hamilton, W.~L., Li{\`o}, P., Bengio, Y.,
  and Hjelm, R.~D. (2018).
\newblock Deep graph infomax.
\newblock {\em arXiv preprint arXiv:1809.10341}.

\bibitem[Vinyals et~al., 2015]{vinyals2015show}
Vinyals, O., Toshev, A., Bengio, S., and Erhan, D. (2015).
\newblock Show and tell: A neural image caption generator.
\newblock In {\em Proceedings of the IEEE conference on computer vision and
  pattern recognition}, pages 3156--3164.

\bibitem[Wang et~al., 2019]{wang2019balanced}
Wang, T., Zhao, J., Yatskar, M., Chang, K.-W., and Ordonez, V. (2019).
\newblock Balanced datasets are not enough: Estimating and mitigating gender
  bias in deep image representations.
\newblock In {\em Proceedings of the IEEE/CVF International Conference on
  Computer Vision}, pages 5310--5319.

\bibitem[Wang et~al., 2020]{wang2020towards}
Wang, Z., Qinami, K., Karakozis, I.~C., Genova, K., Nair, P., Hata, K., and
  Russakovsky, O. (2020).
\newblock Towards fairness in visual recognition: Effective strategies for bias
  mitigation.
\newblock In {\em Proceedings of the IEEE/CVF conference on computer vision and
  pattern recognition}, pages 8919--8928.

\bibitem[Wu et~al., 2020]{wu2020mutual}
Wu, M., Zhuang, C., Mosse, M., Yamins, D., and Goodman, N. (2020).
\newblock On mutual information in contrastive learning for visual
  representations.
\newblock {\em arXiv preprint arXiv:2005.13149}.

\bibitem[Young et~al., 2014]{young2014image}
Young, P., Lai, A., Hodosh, M., and Hockenmaier, J. (2014).
\newblock From image descriptions to visual denotations: New similarity metrics
  for semantic inference over event descriptions.
\newblock {\em Transactions of the Association for Computational Linguistics},
  2:67--78.

\bibitem[Yu et~al., 2016]{yu2016modeling}
Yu, L., Poirson, P., Yang, S., Berg, A.~C., and Berg, T.~L. (2016).
\newblock Modeling context in referring expressions.
\newblock In {\em European Conference on Computer Vision}, pages 69--85.
  Springer.

\bibitem[Zhang et~al., 2019]{zhang2019lookahead}
Zhang, M.~R., Lucas, J., Hinton, G., and Ba, J. (2019).
\newblock Lookahead optimizer: k steps forward, 1 step back.
\newblock {\em arXiv preprint arXiv:1907.08610}.

\bibitem[Zhao et~al., 2017]{zhao2017men}
Zhao, J., Wang, T., Yatskar, M., Ordonez, V., and Chang, K.-W. (2017).
\newblock Men also like shopping: Reducing gender bias amplification using
  corpus-level constraints.
\newblock {\em arXiv preprint arXiv:1707.09457}.

\bibitem[Zhu et~al., 2016]{zhu2016visual7w}
Zhu, Y., Groth, O., Bernstein, M., and Fei-Fei, L. (2016).
\newblock Visual7w: Grounded question answering in images.
\newblock In {\em Proceedings of the IEEE conference on computer vision and
  pattern recognition}, pages 4995--5004.

\end{thebibliography}

%%%%%%%%% APPENDIX
\appendix
\clearpage 
\newpage 
\section{Appendix}
This appendix is organized as follows:
\begin{itemize}
    % \item \textbf{Figure \ref{fig:pseudo}}. Pytorch-style pseudo code
    % \item \textbf{A.1.} Limitations and Broader Impacts
    \item \textbf{A.1.} Discussion on the JSD-based lower bound on MI
    \item \textbf{A.2.} Comparison with Self-Supervised and other pretraining methods
    % \item \textbf{A.3.} Transfer Learning with Backbone Finetuning
    \item \textbf{A.3.} Mutual Information Discriminator
    \item \textbf{A.4.} Ablations on (1) batch sizes, (2) visual encoders, (3) textual encoders, (3) Zero-shot templates
    \item \textbf{A.5.} Training CLIP on the COCO-Captions dataset
\end{itemize}

\subsection{Discussion on JSD-based lower bound on Mutual Information}
Recall that for given random variables $y$ and $z$, their mutual information is defined as a Kullback-Leibler (KL) divergence between their joint distribution $p(y,z)$ and the product of their marginal distributions, $p(y)p(z)$ as, $I(y; z) = D_{\text{KL}}(p(y, z) \, || \, p(y)p(z))$. The above formulation of MI gives rise to the commonly used contrastive objective InfoNCE~\citep{oord2018representation}. Alternatively, the KL-divergence can be replaced with the Jensen-Shannon divergence (JSD) between the joint and the product of marginals as an estimate of the Pointwise Mutual Information(PMI) between two views of the data i.e. $I^{JSD}(y; z) = D_{\text{JSD}}(p(y, z) \, || \, p(y)p(z))$. And as discussed in~\cite{hjelm2018learning}, this formulation of MI leads to the following relation,

\begin{equation}
\begin{split}
    JSD(p(y, z)||p(y) & p(z))  \propto \\& 
    \EE_{y \sim p(y)} [ \EE_{z \sim p(z|y)} [ \log \tfrac{p(z|y)}{p(z)} \\  & - (1 + \tfrac{p(z)}{p(z|y)}) \log \left( 1 + \tfrac{p(z|y)}{p(z)} \right)  ] ] \\
\end{split}
\end{equation}

Now, the quantity inside the expectation above is a concave, monotonically increasing function of the ratio $p(z|y)/p(z)$, which is exactly the exponential of the Pointwise Mutual Information, i.e. $e^{PMI(y,z)}$.

\subsection{Comparison with SSL Pretraining Methods}
In this section, we evaluate the performance of our method against other pre-training frameworks and image-only SSL methods. We observe that CLIP-Lite is comparable or better to image-only SSL learning models trained on downstream ImageNet classification with a frozen ResNet-50 backbone, even though our method is trained on much fewer images, albeit with textual supervision.

% Backbone finetuning ECCV table
\begin{table}[t]
\setlength\tabcolsep{1.7pt}
\renewcommand{\arraystretch}{1.0}
\caption{CLIP-Lite outperforms CLIP-COCO on both VOC and ImageNet classification tasks, and performs comparably to VirTex. CLIP-Lite's performance is comparable or superior to both supervised and self-supervised learning models trained with images alone, even those trained with 10x more images. (IN-Sup. = ImageNet-supervised.)}
\label{tab:transferfull}
\small
  \centering
    \resizebox{\columnwidth}{!}{%
    \begin{tabular}{lcccc}
        \toprule
        \textbf{Method} & \textbf{\# images} & \textbf{Annotations} & \textbf{VOC07} & \textbf{IN-1k} \\ \midrule
        COCO-Sup.      & 118K                & labels            & 86.2           & 46.4           \\
        IN-Sup.         & 1.28M               & labels             &  87.6          & 75.6          \\
        \midrule
        MoCo-COCO       & 118K               & self-sup.            & 67.5           & 46.5           \\
        MoCo-IN v1      & 1.28M              & self-sup.            & 79.4           & 60.8           \\
        PCL v1          & 1.28M              & self-sup.            & 83.1           & 61.5           \\
        SwAV (200 ep.)  & 1.28M              & self-sup.            & 87.9           & 72.7           \\ \midrule
        ICMLM           & 118K               & captions             & 87.5           & 47.9           \\
        VirTex          & 118K               & captions             & 88.7           & 53.8           \\\midrule
        CLIP-COCO       & 118K               & captions             & 74.2           & 33.2              \\ 
        CLIP-Lite       & 118K               & captions             & 88.2           & 55.3           \\ \bottomrule
    \end{tabular}}
\end{table}

% Subsection
\subsection{Mutual Information Discriminator}
As described in main paper, our JSD-based lower-bound on mutual information relies on a discriminator function, $T_{\omega}: \mathcal{Y} \times \mathcal{Z} \to \mathbb{R}$, which distinguishes between samples extracted from the joint distribution, $P(Y, Z)$~i.e.~a positive image-caption pair and the product of marginals, $P(Y)P(Z)$~i.e.~a negative image-caption pair. This discriminator function can be modelled as an arbitrary neural network with parameters $\omega$ that can be jointly optimized with the encoders during training~\citep{belghazi2018mutual}. In this work, we use a projection and alignment based architecture similar to the one presented in Deep InfoMax~\citep{hjelm2018learning}. 

Given a pair of input one-dimensional representations, both vectors are first projected using a projection module with two linear layers separated by a ReLU and a linear shortcut. A  dot-product of these projections is then computed to get alignment scores. The projection function maps these representations to an aligned cross-modal latent space. Separate projection functions are used for image and text representations. Positive and negative pairs of image-text representations are passed through the discriminator to get respective scores which are then used to estimate and maximize mutual information using our objective. This architecture, in addition to being simple and computationally inexpensive, also offers alignment of the representations into a common cross-modal latent space which uses cosine similarity as the distance metric.

\label{apdx:bs}
\subsection{Ablations}
\paragraph{Batch-size Ablations: }
A salient feature of our pre-training framework is that we use a lower-bound on the mutual information that can be optimized with only one negative sample. This allows us to use much smaller batch-sizes compared to the original CLIP~\citep{radford2021learning} model. In this section, we evaluate the PASCAL VOC classification performance of the visual backbones trained with a batch sizes $64$, $128$, $256$, $512$ and $1024$. These ablations are performed with a $2$-layered BERT model as the text-encoder and a ResNet-$50$ as the image encoder for $200$K iterations. 

\begin{table}[ht]
\renewcommand{\arraystretch}{1.2}
\caption{\textbf{Batch size Ablations:} We show the performance of a ResNet-$50$ trained with CLIP-Lite using varying batch-sizes. We observe that the performance drops marginally with the batch size $512$. Additionally, we can see that the model is able to converge fairly well with the significantly lower batch size of $64$.}
\label{tab:ab_bs}
\small
  \centering
    % \resizebox{\columnwidth}{!}{%
    \begin{tabular}{lc}
        \toprule
        \textbf{Batch Size}     &  \textbf{VOC07} \\
        \midrule
        % \band
        64  & 74.7 \\
        128  & 81.3 \\
        256 & 84.9 \\
        512 & 87.5 \\
        1024 & 87.9 \\
        \bottomrule
    \end{tabular}
    % }
\end{table}

\paragraph{Visual Encoder Ablations: }
In this section, we compare the performance of our pretraining method using a ResNet-$18$, ResNet-$50$, and ResNet-$101$  backbones using the downstream PASCAL VOC classification task. These ablations are performed with a $2$-layered BERT model as the text-encoder with a batch-size of $512$ for $200$K iterations.

\begin{table}[ht]
\renewcommand{\arraystretch}{1.2}
\caption{\textbf{Visual Encoder Ablations:} We show the performance of CLIP-Lite using 3 visual backbones of varying sizes.}
\label{tab:ab_ve}
\small
  \centering
    % \resizebox{\columnwidth}{!}{%
    \begin{tabular}{lc}
        \toprule
        \textbf{Visual Backbone}     &  \textbf{VOC07} \\
        \midrule
        % \band
        ResNet-18  & 83.8 \\
        ResNet-50  & 87.5 \\
        ResNet-101 & 87.8 \\
        \bottomrule
    \end{tabular}
    % }
\end{table}

\paragraph{Text Encoder Ablations: }
In this section, we compare the downstream PASCAL VOC~\citep{everingham2010pascal} classification performance of a ResNet-$50$ visual backbone pretrained using a text encoder transformer with varying capacities. We train 4 transformer variants, $(1)$ pre-trained BERT\textsubscript{base}~\citep{devlin2018bert}, $(2)$ $2$-layered, $(3)$ $4$-layered, $(4)$ $6$-layered, and a $(5)$ $12$-layered BERT-like transformer. These ablations are performed with a ResNet-$50$ as the image encoder with a batch-size of $512$ for $200$K iterations.

\begin{table}[ht]
\renewcommand{\arraystretch}{1.2}
\caption{\textbf{Text Encoder Ablations:} We show the performance of a ResNet-$50$ trained with CLIP-Lite using different text encoders. We observe that the performance drops marginally when training from scratch. Additionally, we also see that using a transformer with $2$-layers works almost as well as a $12$-layered transformer when trained from scratch.}
\label{tab:ab_te}
\small
  \centering
    % \resizebox{\columnwidth}{!}{%
    \begin{tabular}{lc}
        \toprule
        \textbf{Text Encoder}     &  \textbf{VOC07} \\
        \midrule
        % \band
        BERT\textsubscript{base} init.  & 88.1 \\
        $2$-layers  & 87.5 \\
        $4$-layers & 87.6 \\
        $6$-layers & 87.6 \\
        $12$-layers & 87.9 \\
        \bottomrule
    \end{tabular}
    % }
\end{table}

\paragraph{Zero-shot classification templates}
While performing zero-shot classification, we use the class names of target images to generate captions that the images should align with. The performance is compared when captions are generated using three different templates. We test three different class prompt templates and compare our performance against an equivalently trained CLIP model on the COCO dataset. As seen in Table~\ref{tab:zero1}, both CLIP and CLIP-Lite prefer more descriptive prompts. 

\begin{table}[h]
\caption{\textbf{Zero-Shot Templates on CIFAR-10:} We evaluate different prompts and find the CLIP-Lite prefers more descriptive prompts.}
\label{tab:zero1}
\renewcommand{\arraystretch}{1.2}
  \centering
    \resizebox{\columnwidth}{!}{%
    \begin{tabular}{lcc}
        \toprule
        \textbf{Class Prompt}         & \textbf{CLIP-COCO} & \textbf{CLIP-Lite} \\
        \midrule
        ``a \{class name\}''              & 13.3               & 30.8               \\
        ``a picture of a \{class name\}'' & 14.5               & 32.6               \\
        ``a photo of a \{class name\}''   & 16.3               & 33.0               \\
        \bottomrule
    \end{tabular}
    }
\end{table}

\subsection{Training CLIP on COCO-Captions Dataset}
We use a CLIP model trained on the COCO dataset as a baseline for several demonstrated tasks. For this purpose, we use an open-source implementation\footnote{\tiny \url{https://github.com/mlfoundations/open\_clip}} of CLIP. We train a standard ResNet-50~\citep{he2016deep} based CLIP model that takes in a $224\times224$ image and generates $2048$-dimensional features at the pre-logit layer. For textual encoding, we use a transformer~\citep{vaswani2017attention} model and use the output \texttt{[CLS]} token as the text representation. We use the COCO Captions dataset~\citep{chen2015microsoft} which has $118$K images with five captions per image. During training time we apply (1) random cropping, (2) color jittering, (3) random horizontal flips while interchanging the words ‘left’ and ‘right’ in the caption, and (4) normalization using the ImageNet image mean. We train using the Adam Optimizer~\citep{kingma2014adam} with decoupled weight decay regularization~\citep{loshchilov2016sgdr} for all weights except gains or biases. We perform distributed training across $8$ GPUs with batch normalization~\citep{ioffe2015batch} per GPU with an overall batch-size of $1024$. We warm-up to the initial learning rate in $10$K steps and decay to zero with the cosine schedule. We found that using the learning rate of $10^4$ works slightly better ($+1.4\%$ on VOC07) than the originally recommended $5 \times 10^5$. We also found that the performance incrementally improves ($+1.9\%$ on VOC07) with longer training therefore we train for $250$K iterations, similar to ours. All other training details and hyper-parameters were kept the same as the original work~\citep{radford2021learning}. Please note that the ResNet-50 backed CLIP model trained by us on the COCO dataset outperforms (+1.2\% Zero-shot Acc. on CIFAR10) publicly available weights\footnote{\tiny \url{https://github.com/revantteotia/clip-training/blob/main/zero_shot_eval_output/coco_trained_clip_observations.md}}.

\end{document}